\documentclass[times,twocolumn,final]{elsarticle}

\usepackage{medima}
\usepackage{framed,multirow}

\usepackage{amssymb}
\usepackage{latexsym}

\usepackage{url}
\usepackage{xcolor}

\usepackage{hyperref}

\usepackage{amsmath,amssymb,amsfonts,bm}
\usepackage{graphicx}
\usepackage{textcomp}
\usepackage{algorithm,algpseudocode,lipsum}

\usepackage{epstopdf} 
\usepackage{tabularx,threeparttable}
\usepackage{array}
\usepackage{mathtools}
\usepackage{mathrsfs}
\usepackage{booktabs}
\usepackage{multirow}
\usepackage{url}
\usepackage{color}
\usepackage{dsfont}
\usepackage{subcaption}
\usepackage{caption}
\usepackage{subcaption}
\usepackage{hyperref}

\def\ie{\emph{i.e.}}
\def\eg{\emph{e.g.}}

\newcommand\T{\rule{0pt}{2.1ex}}       
\newcolumntype{?}[1]{!{\vrule width #1}}

\definecolor{newcolor}{rgb}{.8,.349,.1}

\journal{Artificial Intelligence in Medicine}

\begin{document}

\verso{Yijun Yang \textit{et~al.}}

\begin{frontmatter}

\title{HCDG: A Hierarchical Consistency Framework for Domain Generalization on Medical Image Segmentation}%

\author[1]{Yijun Yang\fnref{fn1}}
\author[2]{Shujun Wang\fnref{fn1}}
\author[1,3]{Lei Zhu}
\author[4]{Lequan Yu\corref{cor1}}

\address[1]{Robotics and Autonomous Systems Thrust, The Hong Kong University of Science and Technology (Guangzhou), China}
\address[2]{Department of Applied Mathematics and Theoretical Physics, University of Cambridge, United Kingdom}
\address[3]{Department of Electronic and Computer Engineering, The Hong Kong University of Science and Technology, Hong Kong, China}
\address[4]{Department of Statistics and Actuarial Science, The University of Hong Kong, Hong Kong, China}

\cortext[cor1]{Corresponding author.}
\ead{lqyu@hku.hk}
\fntext[fn1]{This is author footnote for co-first authors.}


\begin{abstract}
Modern deep neural networks struggle to transfer knowledge and generalize across diverse domains when deployed to real-world applications. 
Currently, domain generalization (DG) is introduced to learn a universal representation from multiple domains to improve the network generalization ability on unseen domains. 
However, previous DG methods only focus on the data-level consistency scheme without considering the synergistic regularization among different consistency schemes. 
In this paper, we present a novel Hierarchical Consistency framework for Domain Generalization (HCDG) by integrating Extrinsic Consistency and Intrinsic Consistency synergistically.
Particularly, for the Extrinsic Consistency, we leverage the knowledge across multiple source domains to enforce data-level consistency.
To better enhance such consistency, we design a novel Amplitude Gaussian-mixing strategy into Fourier-based data augmentation called DomainUp.
For the Intrinsic Consistency, we perform task-level consistency for the same instance under the dual-task scenario.
We evaluate the proposed HCDG framework on two medical image segmentation tasks, \ie, optic cup/disc segmentation on fundus images and prostate MRI segmentation. 
Extensive experimental results manifest the effectiveness and versatility of our HCDG framework. 

\end{abstract}


\end{frontmatter}


\section{Introduction}

Deep neural networks (DNNs) have demonstrated advanced progress on diverse medical image analysis tasks~\citep{piantadosi2020multi,calisto2021emonas,survarachakan2022deep}. 
Most of these achievements depend on the special requirement that networks are trained and tested on the samples drawn from the same distribution or domain. 
Once such requirement fails, \ie, domain shift~\citep{domainshift2012} exists, networks are very likely to generate unsatisfied performance due to the limited generalization ability.
Typically, in the real clinical setting, medical images are usually captured by different institutions with various types of scanners vendors, patient populations, in the field of view and appearance discrepancy~\citep{ting2019artificial}, which makes learned models struggle to transfer knowledge and generalize across these institutions. 
Since training the specific model for each medical center is unrealistic and laborious, it is necessary to enhance the deep model generalization ability across different and even new clinical sites.

The community has attacked domain shift bottleneck so far mainly in two directions.
Firstly, Unsupervised Domain Adaptation (UDA) exploits prior knowledge extracted from unlabeled target domain images to achieve model adaptation. 
Although UDA-based approaches could avoid time-consuming annotations from the target domain, collecting target images in advance is hard to meet in practice.
This inspires another direction, Domain Generalization (DG), which aims to learn a universal representation from multiple source domains without any target domain information.
In this paper, we intend to utilize the DG-based method to improve the network generalization on medical image segmentation tasks.

Under the domain generalization scope, data manipulation~\citep{tobin2017domain,zhou2020deep},
domain-invariant representation learning~\citep{Muandet2013,ganin2015unsupervised,arjovsky2019invariant} and meta-learning techniques~\citep{li2018learning,dou2019domain} have achieved remarkable success.
Meanwhile, consistency regularization, which prevails in Semi-Supervised Learning (SSL) and UDA, has been introduced to mitigate performance degradation by forcing the model to learn invariant information from perturbed samples and has shown promising results in different tasks.
However, most previous consistency regularization-based works~\citep{pyramid2019,fourier2021} simply enforce data-level consistency by generating new domains with novel appearance and then minimizing the discrepancy between original and generated domains for the same instance. 
Such consistency, based on data-level perturbation, usually requires extrinsic knowledge from other source domains and heavily depends on the quality of generated domains.
To overcome the above shortcomings, we are motivated to explore other kinds of consistency regularization for the DG problem. 
Particularly, based on the observation that related tasks inherently introduce prediction perturbation during the network training, we may leverage the intrinsic consistency of related tasks from the same instance without extrinsic generated domains to encourage the network to learn generalizable representation.
Additionally, how to integrate data-level consistency and task-level consistency and leverage their complementary effect is also well worth exploring.

To this end, we present a novel Hierarchical Consistency framework for Domain Generalization (HCDG) by harnessing Extrinsic and Intrinsic Consistency simultaneously.
%
To the best of our knowledge, we are the first to introduce task-level perturbation into DG and integrate several kinds of consistency regularization into a hierarchical cohort, enforcing the smoothness assumption into both input space and output space.
For Extrinsic Consistency, we leverage the knowledge across multiple source domains to deploy data-level consistency.
Inspired by observation that the phase and amplitude components in the Fourier spectrum of signals retain the high-level semantics (\eg, structure) and low-level statistics (\eg, appearance, color), we introduce an improved Fourier-based Amplitude Gaussian-mixing (AG) method called DomainUp, to produce augmented domains with richer variability compared with the previous Amplitude Mix (AM) scheme~\citep{fourier2021}. 
For Intrinsic Consistency, we perform task-level consistency for the same instance under two related tasks: image segmentation and boundary regression.
The Extrinsic and Intrinsic Consistency are further integrated into a teacher-student-like cohort to facilitate network learning.
We evaluate the proposed HCDG on two medical image segmentation tasks, \ie, optic cup/disc segmentation on fundus images and prostate segmentation on MRI images. Our HCDG framework achieves state-of-the-art performance in both tasks. 
%

The main contributions are summarized as follows.
\begin{enumerate}
    \item[(i)] 
    We develop an effective HCDG framework for generalizable medical image segmentation by simultaneously integrating Extrinsic and Intrinsic Consistency. 
    \item[(ii)] We design a novel Amplitude Gaussian-mixing strategy for Fourier-based data augmentation by introducing pixel-wise perturbation in the amplitude spectrum to highlight core semantic structures.
    \item[(iii)] Extensive experiments on two medical image segmentation benchmark datasets validate the efficacy and universality of the framework and HCDG clearly outperforms many state-of-the-art DG methods. Code is in \url{https://github.com/scott-yjyang/HCDG}.

\end{enumerate}

\section{Related Work}
\label{sec:relatedwork}

We first summarize recent works on medical image segmentation tasks, especially retinal fundus segmentation and prostate MRI segmentation.
Then, we review related techniques of domain generalization, which enhance the generalization ability of deep networks. Finally, the development of consistency regularization is elaborately introduced.

\subsection{Medical Image Segmentation}
DNNs have been widespread in medical image segmentation tasks, such as cardiac segmentation from MRI~\citep{yu2017automatic}, organ segmentation from CT~\citep{zhou2017fixed,li2018h}, and skin lesion segmentation from dermoscopic images~\citep{yuan2017improving}.  
In this paper, we mainly focus on two medical tasks, \ie, OC/OD segmentation from retinal fundus images~\citep{beal2019}, and the prostate segmentation from T2-weighted MRI~\citep{mridataset2020}.
Previously, \cite{fu2018joint} and \cite{wang2019patch} showed competing results on jointly segmenting OC/OD, while \cite{milletari2016v} and \cite{yu2017volumetric} successfully improved the performance on prostate segmentation.
However, most of the methods lack the generalization ability and tend to generate high test errors on unseen target datasets.
Thus, a more generalizable method is highly desired to alleviate performance degradation.

\begin{figure*}[t]
\centering
\includegraphics[width=0.9\textwidth]{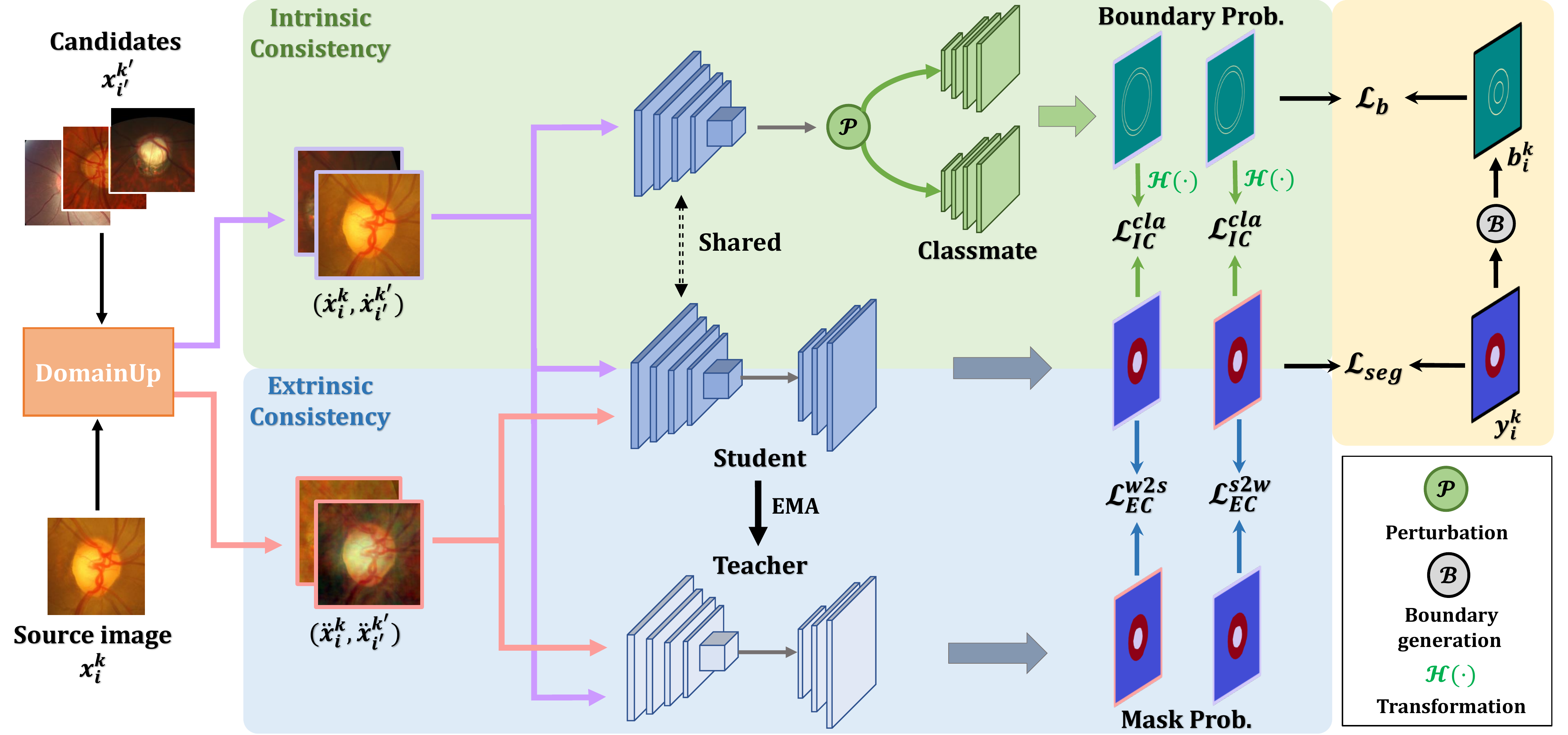} 
\caption{Overview of the proposed HCDG framework. Firstly, the weakly and strongly augmented replicas $(\dot{x}_i^k,\dot{x}_{i'}^{k'})$,$(\ddot{x}_i^k,\ddot{x}_{i'}^{k'})$ are generated by DomainUp from the source image $x_i^k$ and candidates $x_{i'}^{k'}$. For Extrinsic Consistency, both replicas are sent to the student and teacher model to conduct image segmentation task. They leverage the additional knowledge from 
interpolated domains to enforce data-level consistency. For Intrinsic Consistency, only the weakly augmented replica is fed into the incorporated Classmates module to conduct the dual-task, \ie, boundary regression task.}
\label{fig:framework}
\end{figure*}

\subsection{Domain Generalization}
Domain generalization aims to learn a general model from multiple source domains such that the model can directly generalize to arbitrary unseen target domains. 
Recently, many DG approaches have achieved remarkable results.
Early DG works mainly follow the representation learning spirit via kernel methods~\citep{Muandet2013,li2018domain}, domain adversarial learning~\citep{ganin2015unsupervised,li2018adversarial}, invariant risk minimization~\citep{arjovsky2019invariant,guo2021out}, multi-component analysis~\citep{zunino2021explainable}, and generative modeling~\citep{qiao2020learning}.
Data manipulation is one of the cheapest ways to tackle the dearth of training data and enhance the generalization capability of the model by two popular techniques: data generation and data augmentation. 
For example, domain randomization~\citep{tobin2017domain}, transformation network trained adversarially~\citep{zhou2020deep}, and Mixup~\citep{zhang2017mixup} are utilized to generate more training samples.
Meanwhile, \cite{fourier2021} introduced the Fourier-based data augmentation for DG by linearly distorting the amplitude information.
DG has also been studied in general machine learning paradigms.
\cite{li2018learning} designed a model agnostic training procedure, which is derived from meta-learning. \cite{carlucci2019domain} formulated a self-supervision task of solving jigsaw puzzles to learn generalized representations. 
Inspired by Lottery Ticket Hypothesis~\citep{frankle2018lottery}, \cite{cai2021generalizing} proposed to learn domain-invariant parameters of the model during training.
Different from all the above methods, our work provides a novel Hierarchical Consistency-based perspective for DG. By integrating Extrinsic Consistency and Intrinsic Consistency, our approach outperforms many current DG methods.

\subsection{Consistency Regularization}
The consistency regularization is widely used in supervised and semi-supervised learning but has not played a significant role in DG yet.
\cite{sajjadi2016regularization} first introduced a consistency loss to utilize the stochastic nature of data augmentation and minimize the discrepancy between the predictions of multiple passes of a training sample through the network.  \cite{meanteacher2017} designed a teacher-student model to provide better consistency alignment.
Besides, \cite{pyramid2019} proposed a pyramid consistency to learn a model with high generalizability via domain randomization, which still rests on data-level perturbation. 
Very recently, task-level consistency has been used for semi-supervised learning~\citep{dualtask2021}.
To our best knowledge, there is no work exploring task-level consistency for DG problem.

\section{Methodology}
Fig.~\ref{fig:framework} depicts the proposed Hierarchical Consistency framework for Domain Generalization (HCDG).
We consider a training set of multiple source domains $\mathcal{S} = \{S_1,...,S_K\}$ with $N_k$ labeled samples $\{(x_i^k, y_i^k)\}_{i=1}^{N_k}$ in the $k$-th domain $S_k$, where $x_i^k$ and $y_i^k$ denote the images and labels, respectively.
Our HCDG framework learns a domain-agnostic model $f_\theta: X\rightarrow Y$ using $K$ distributed source domains by Extrinsic Consistency and Intrinsic Consistency simultaneously, so that it can directly generalize to a completely unseen domain $\mathcal{T}$ with mitigating performance degradation. 
\subsection{Extrinsic Consistency (EC)}
We first exploit consistency regularization from the extrinsic aspect, \ie, leveraging extra knowledge from other source domains, to enforce data-level consistency for each instance.
Specifically, we propose a new paradigm of Fourier-based data augmentation, named Amplitude Gaussian-mixing (AG), to perturb the spectral amplitude and then generate augmented images. 
Based on AG, we further design a DomainUp scheme to provide image from new domains with enough variability.
Finally, we utilize a mean-teacher framework to minimize the discrepancy between augmented replicas from the same instance with dual-view consistency regularization. 

\begin{figure}[t]
\centering
\includegraphics[width=1.0\columnwidth]{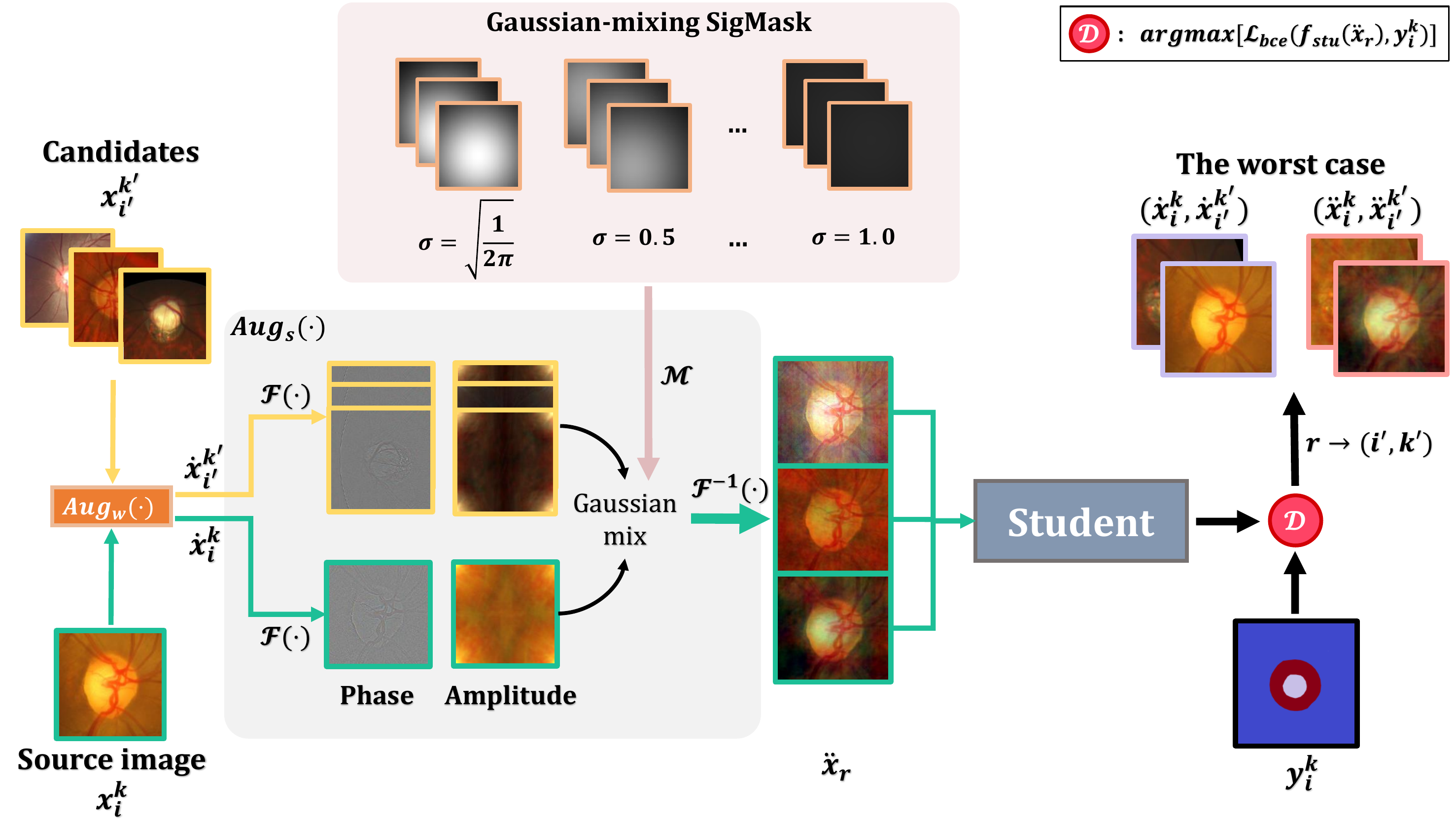} 
\caption{Illustration of DomainUp. For each source image $x^k_i$, DomainUp conducts weakly data augmentation $Aug_w(\cdot)$ and AG-based strongly data augmentation $Aug_s(\cdot)$ to get weakly augmented $\dot{x}_i^k$ and strongly augmented $\ddot{x}_r$. The worst augmented case is then selected by the maximal segmentation loss from $\ddot{x}_r$.}

\label{fig:domainup}

\end{figure}

\subsubsection{Amplitude Gaussian-mixing}
Previous Fourier-based augmentation work~\citep{fourier2021} linearly mixes the spectral amplitude of the whole image and keeps the phase information invariant to synthesize interpolated domains. 
This can promise that the structure of object does not change while the amplitude perturbation makes the model more robust to different appearances without affecting segmentation results.
However, this strategy treats each pixel equally during mixing, which can hardly distinguish the magnitude of semantic structures in the center and marginal areas.
Thus, we design a novel Fourier-based Amplitude Gaussian-mixing (AG) strategy for amplitude perturbation by introducing a significance mask, SigMask $\mathcal{M}$, for linear interpolation, where the Gaussian-like $\mathcal{M}$ is used to control the perturbation magnitude at each pixel.
Specifically, we first extract the frequency space signal $\mathcal{F}(x_i^k)$ of sample $x_i^k \in \mathbb{R}^{H\times W\times C}$ through fast Fourier transform~\citep{nussbaumer1981fast} and further decompose $\mathcal{F}(x_i^k)$ into an amplitude spectrum $\mathcal{A}(x_i^k)\in \mathbb{R}^{H\times W\times C}$ and a phase spectrum $\mathcal{P}(x_i^k)\in \mathbb{R}^{H\times W\times C}$.
For each $x_i^k$, its perturbed amplitude spectrum $\hat{\mathcal{A}}(x_i^k)$ is calculated according to $\mathcal{A}(x_i^k)$ and  $\mathcal{A}(x_{i^{'}}^{k^{'}})$ of counterpart sample $x_{i^{'}}^{k^{'}}$ via $\mathcal{M}$ following
\begin{align}
    \hat{\mathcal{A}}(x_i^k) = (1-\mathcal{M})\odot \mathcal{A}(x_i^k) + \mathcal{M}\odot \mathcal{A}(x_{i^{'}}^{k^{'}}),\label{eq:amplitude_perturbation}
\end{align}
where $\odot$ denotes element-wise multiplication.
Then, we generate the augmented image of interpolated domain via inverse Fourier Transform $\mathcal{F}^{-1}(\hat{\mathcal{A}}(x_i^k),\mathcal{P}(x_i^k))$.

The values of SigMask $\mathcal{M}$ follow a Gaussian distribution and the value of each pixel $\mathcal{M}(a,b)$ is computed  with 

\begin{equation}
\begin{aligned}
    \mathcal{M}(a,b)=\frac{1}{2\pi\sigma^2}e^{-\frac{(a-\mu_1)^2+(b-\mu_2)^2}{2\sigma^2}},\\
\end{aligned}
\label{eq:gaussian}
\end{equation}
where $\sigma \sim U(\frac{1}{\sqrt{2\pi}}, \eta)$ and $\mu_1,\mu_2 \sim U(-\frac{t}{2}, \frac{t}{2})$.
It is worth noting that we scale the range of $a, b$ to $[-t, t]$ to avoid pretty small significances in the marginal area of SigMask. 
The variance $\sigma^2$ controls the peak value of the above Gaussian-mixing function under fixed $t$, while the mean $\mu_1, \mu_2$ decide the position of the peak. 
To prevent generating outliers, we deduce the lower bound of $\sigma$ to $\frac{1}{\sqrt{2\pi}}$ to promise $\mathcal{M}_{max}\leqslant 1$.
We also control the upper bound via the hyper-parameter $\eta$.

Our AG is simple but effective bringing three benefits: (1) the Gaussian-mixing function highlights the core information by endowing the center area of the image with more magnitude than the marginal area;
(2) we can generate adaptive center areas
by empirically sampling $\mu_1, \mu_2$ from $U(-\frac{t}{2}, \frac{t}{2})$ to cope with uncertain positions of core semantics;
(3) stability has been improved by controlling the variance of the Gaussian-mixing function instead of directly changing $\mathcal{M}(a,b)$.
Mention that our AG is robust to different datasets and does not need to elaborately tune the hyper-parameters.

\subsubsection{DomainUp}
To obtain more informative interpolated samples, for each source image $x^k_i$, we search for the worst augmented case from $\omega \times K$ candidates, where $\omega$ is the instance number sampled from each source domain. 
As shown in Fig.~\ref{fig:domainup}, we first obtain the weakly augmented source image $\dot{x}_i^k$ and candidates $\dot{x}_{i^{'}}^{k^{'}}$ by standard augmentation protocol $Aug_w(\cdot)$ (\eg, random scaling and flipping), and then deploy the Fourier-based AG augmentation $Aug_s(\cdot)$ on $\dot{x}_i^k$ and each $\dot{x}_{i^{'}}^{k^{'}}$ to acquire the corresponding strongly augmented replica $\ddot{x}_r$.
The worst augmented case can be selected by the maximal supervised loss value from $\ddot{x}_r$.
We combine the maximal segmentation loss $\mathcal{L}_{seg}^{s}$ from $\ddot{x}_r$ and normal segmentation loss $\mathcal{L}_{seg}^{w}$ from $\dot{x}_i^k$ as the total supervised segmentation loss $\mathcal{L}_{seg}$ following

\begin{equation}
    \begin{aligned}
    \mathcal{L}_{seg}^{s} &= \mathop{max}\limits_{r \in [\omega \times K]}
    \mathcal{L}_{bce}(f_{stu}(\ddot{x}_r),y_i^k),\\
    \mathcal{L}_{seg}^{w} &= \mathcal{L}_{bce}(f_{stu}(\dot{x}_i^k),y_i^k),\\
    \mathcal{L}_{seg} &= \mathcal{L}_{seg}^{s} + \mathcal{L}_{seg}^{w},
    \end{aligned}
\label{eq:segloss}
\end{equation}
where $f_{stu}(\cdot)$ is the prediction of the student model.
%


\subsubsection{Dual-view Consistency Regularization}
After acquiring weakly and strongly augmented images, we explicitly enforce a dual-view consistency regularization for Extrinsic Consistency. 
Specifically, the consistency is implemented with a momentum-updated teacher model to provide dual-view instance alignment. 
We feed the weakly and strongly augmented images into both student and teacher networks (with the same architecture) and then minimize their network output discrepancy with the Kullback-Leibler (KL) divergence: 
\begin{align}
    \mathcal{L}_{EC}^{w2s} &= KL(\phi(f_{stu}(\dot{x}_i^k)/T) || \phi(f_{tea}(\ddot{x}_i^k)/T)), \nonumber\\
    \mathcal{L}_{EC}^{s2w} &= KL(\phi(f_{stu}(\ddot{x}_i^k)/T) || \phi(f_{tea}(\dot{x}_i^k)/T)), \label{eq:ecs2w}
\end{align}
where $f_{tea}(\cdot)$ is the teacher model prediction, $\phi(\cdot)$ denotes the softmax operation, and $T$ represents the temperature~\cite{temperature2015} to soften the outputs.
Overall, the objective function of EC is composed of the supervised segmentation loss and the consistency KL loss 
as $\mathcal{L}_{EC} = \mathcal{L}_{seg} + \gamma(\mathcal{L}_{EC}^{w2s} + \mathcal{L}_{EC}^{s2w})$ with a balancing hyper-parameter $\gamma$.


\subsection{Intrinsic Consistency (IC)}
Different from EC, IC introduces a Classmate module for task-level consistency guided by intrinsic perturbation. 
As the corresponding predictions of related tasks for the same sample in the output space have inherent difference, consistency can be enforced without extra knowledge after proper transformation.

\subsubsection{Classmate Module} 
To conduct dual tasks, we incorporate a new Classmate Module with $Q$ decoders into the student model.
We then achieve task-level constraints between predictions of two tasks after transformation.
To enrich the learned representations of the model, we also apply a feature-level perturbation~$\mathcal{P}(\cdot)$ on the feature map $z_i^k$ (\ie, feature dropout or noise), making our IC more powerful.
Specifically, each classmate is composed of four convolutional layers followed by ReLU and batch normalization layers, and receives the perturbed feature map of the weakly augmented image $\dot{x}_i^k$. 
In practice, we define boundary regression as the second task to capture geometric structure.
To generate the boundary ground truth $b_i^k$, we apply level set function~\cite{li2005level} and normalization onto the mask ground truth. 
The supervised boundary loss is formulated with mean square error (MSE)
\begin{equation}
    \begin{aligned}
    \mathcal{L}_b &= \sum_{j=1}^{Q}\frac{1}{N}\sum_{{z}_i^k}(b_i^k-tanh(g_j(\mathcal{P}{({z}_i^k)})))^2,
    \end{aligned}
\label{eq:lossbou}
\end{equation}
where $N$ is the number of the labeled samples, $g_j(\cdot)$ is the prediction of the classmate $j$ and $tanh(\cdot)$ is the tanh activation function to normalize the output into $[-1, 1]$. 
We set the number of classmates $Q=2$ to balance the accuracy and efficiency.
Note that while each classmate receives a different version of the perturbed feature map, all classmates are required to generate consistent predictions with the student decoder under both task-level perturbation and feature-level perturbation.

\subsubsection{Dual-classmate Consistency Regularization}
We define $\mathcal{O}_{in}, \mathcal{O}_{out}$ as the inside and outside area of the target object. 
Since the level set function generates the boundary ground truth as the signed distance map, we strive to transform the regressed boundary into the map with 1 in $\mathcal{O}_{in}$ and 0 in $\mathcal{O}_{out}$ similar to the mask ground truth.
Consequently, we scale the regressed boundary prediction followed by a sigmoid-like function to approximate the predicted mask, \ie, the smooth Heaviside function~\citep{sigmoid2020}:
\begin{equation}
    \begin{aligned}
    \mathcal{H}(x) &= 1/({1+e^{-\delta\cdot x}}),
    \end{aligned}
\label{eq:transformation}
\end{equation}
where $\delta$ denotes a scaling factor.
The approximate transformation function maps the prediction space of boundary to that of mask while still maintaining the task-level diversity. 
Thus, task-level consistency can be enforced between the mask prediction $f_{stu}(\dot{x}_i^k)$ and the transformed boundary prediction $\mathcal{H}(g_j(\mathcal{P}{({z}_i^k)}))$ from the same sample by exploiting intrinsic knowledge:
\begin{equation}
    \begin{aligned}
    \mathcal{L}_{IC}^{cla} = \sum_{j=1}^{Q} KL(\phi(f_{stu}(\dot{x}_i^k))||\phi(\mathcal{H}(g_j(\mathcal{P}{({z}_i^k)})))). 
    \end{aligned}
\label{eq:iccla}
\end{equation}
Then, the objective of IC consists of the boundary MSE loss and the consistency KL loss as $\mathcal{L}_{IC} = \mathcal{L}_b + \gamma\mathcal{L}_{IC}^{cla}, $
where $\gamma$ is empirically the same as that of EC. 
Overall, the total objective function of training the framework is summed up as 
\begin{equation}
    \mathcal{L}_{total} = \mathcal{L}_{EC} + \mathcal{L}_{IC}. 
    \label{totalloss}
\end{equation}

\begin{algorithm}[!t]
\caption{Training procedure of our HCDG framework} 
\label{alg:1}
{\bf Input:} 
A mini-batch of $(x_i^k, y_i^k)$ from source domains $\mathcal{S}$.\\
{\bf Output:} 
The prediction $p_i^k$ of input $\hat{x}_i^k$ from unseen target domain $\mathcal{T}$.
\begin{algorithmic}[1]
\State $\theta_{stu}=\{\theta_{stu}^{e},\theta_{stu}^{d}\}, \theta_{tea}, \theta_{cla}$ $\leftarrow$ initialize

\While{not converge} 
    \State $(x_i^k, y_i^k)$ sampled from source domains $\mathcal{S}$
    \State $(x_{i^{'}}^{k^{'}}, y_{i^{'}}^{k^{'}})$ sampled from source domains $\mathcal{S}$
    \State DomainUp generates $\dot{x}_i^k, \ddot{x}_i^k, \dot{x}_{i^{'}}^{k^{'}}, \ddot{x}_{i^{'}}^{k^{'}}$
    \State $\dot{x}_i^k = concat(\dot{x}_i^k,\dot{x}_{i^{'}}^{k^{'}})$, $\ddot{x}_i^k = concat(\ddot{x}_i^k,\ddot{x}_{i^{'}}^{k^{'}})$
    \State $y_i^k = concat(y_i^k, y_{i^{'}}^{k^{'}})$
    \State Generate boundary ground truth $b_i^k$ from $y_i^k$ 
    \State Calculate $\mathcal{L}_{seg}$ as Eq.~\eqref{eq:segloss}
    \State Update $\theta_{stu} \stackrel{+}{\leftarrow}-\Delta_{\theta_{stu}}\mathcal{L}_{seg}$
    \State Calculate $\mathcal{L}_{EC}^{w2s}$ and $\mathcal{L}_{EC}^{s2w}$ as Eq.~\eqref{eq:ecs2w}
    \State Update $\theta_{stu} \stackrel{+}{\leftarrow}-\Delta_{\theta_{stu}}\gamma(\mathcal{L}_{EC}^{w2s}+\mathcal{L}_{EC}^{s2w})$
    \State Calculate $\mathcal{L}_b$ as Eq.~\eqref{eq:lossbou}
    \State Update $\theta_{cla} \stackrel{+}{\leftarrow}-\Delta_{\theta_{cla}}\mathcal{L}_{b}$
    \State Update $\theta_{stu}^{e} \stackrel{+}{\leftarrow}-\Delta_{\theta_{stu}^{e}}\mathcal{L}_{b}$
    \State Calculate $\mathcal{L}_{IC}^{cla}$ as Eq.~\eqref{eq:iccla}
    \State Update $\theta_{cla} \stackrel{+}{\leftarrow}-\Delta_{\theta_{cla}}\gamma\mathcal{L}_{IC}^{cla}$
    \State Update $\theta_{stu} \stackrel{+}{\leftarrow}-\Delta_{\theta_{stu}}\gamma\mathcal{L}_{IC}^{cla}$
    \State Momentum Update $\theta_{tea}$ via EMA as Eq.~\eqref{eq:ema}
\EndWhile

\State Calculate $p_i^k=f_{stu}(\hat{x}_i^k)$
\State \Return $p_i^k$
\end{algorithmic}
\end{algorithm}

\subsection{Training Strategy}

In our framework, the architecture of the teacher model is identical to that of the student model, while classmates as extra decoders share the encoder with the student model. 
Rather than gradients flowing through the teacher model during backpropagation, the teacher model receives parameters from the student model via exponential moving average (EMA) following the previous mean-teacher framework~\citep{meanteacher2017}:
\begin{equation}
    \begin{aligned}
    \theta_{tea} = m\cdot\theta_{tea} + (1-m)\cdot\theta_{stu}, 
    \end{aligned}
\label{eq:ema}
\end{equation}
where $m$ is the decay rate during the updating. 
The classmate module does not engage in EMA.

The encoder of the student model is simultaneously updated by gradient flows from image segmentation task and boundary regression task.
Hence, it possesses a stronger ability to recognize the active structure and boundary of objects. 
In the testing phase, we simply use the student model for fair comparison. 
The detailed training procedure is illustrated in Algorithm~\ref{alg:1}. 
Note that we enforce the dataloader to sample a mini-batch which includes all source domain images for better training.

\section{Experiments}

\begin{table*} [!ht]
	\centering
	\caption{Statistics of the public Fundus and Prostate MRI datasets in our experiments.}
	\label{tab:datasetstatistics}
	\begin{tabular}
		{c|c|l|c|l}
		\toprule[1pt]
		\textbf{Tasks} &    \textbf{Domain No.} & \textbf{Dataset} &   T\textbf{raining samples(test)}  & \textbf{Scanners (Institutions)}   \\
		\hline
		\multirow{4}{*}{{Fundus}} & Domain 1 & Drishti-GS~\citep{sivaswamy2015comprehensive}   &  50(51)  & (Aravind eye hospital)  \\
		\cline{2-5} & Domain 2 & RIM-ONE-r3~\citep{fumero2011rim}  &99(60) & Nidek  AFC-210 \\
		\cline{2-5} &Domain 3 & REFUGE  (train)~\citep{orlando2020refuge} &  320(80)  & Zeiss Visucam 500  \\
		\cline{2-5} &Domain 4 & REFUGE (val)~\citep{orlando2020refuge} &  320(80)   & Canon CR-2  \\
		\toprule[1pt]
		
		        \hline
		\multirow{6}{*}{{Prostate MRI}} & {Domain 1} & {NCI-ISBI 2013}~\citep{bloch2015nci}  &  {30} &{(RUNMC)}    \\
		\cline{2-5} & {Domain 2} & {NCI-ISBI 2013}~\citep{bloch2015nci} &{30} & {(BMC)} \\
		\cline{2-5} &{Domain 3} & {I2CVB}~\citep{lemaitre2015computer} &  {19} &  {(HCRUDB)}  \\
		\cline{2-5} &{Domain 4} & {PROMISE12}~\citep{litjens2014evaluation} &  {13}  & {(UCL)} \\
		\cline{2-5} &{Domain 5} & {PROMISE12}~\citep{litjens2014evaluation} &  {12} &  {(BIDMC)}  \\
		\cline{2-5} &{Domain 6} & {PROMISE12}~\citep{litjens2014evaluation} &  {12}  & {(HK)} \\
		\toprule[1pt]
	\end{tabular}
\end{table*}

\begin{table*}[!tbp]
    \renewcommand\arraystretch{1.2}
    \centering
        \caption{\small{Comparison with recent domain generalization methods on the OC/OD segmentation task. The top two values are emphasized using \textbf{bold} and \underline{underline}, respectively.}}
        \label{tab:comparisonsfundus}
        \vspace{-2mm}
        \resizebox{1.0\textwidth}{!}{%
        \setlength\tabcolsep{3.0pt}
        \scalebox{0.6}{
        \begin{tabular}{c|c  c  c  c c| c cccc |c||c  c  c  c c| c cccc |c}
            \hline
            \hline
             Task&\multicolumn{5}{c|}{Optic Cup Segmentation} &\multicolumn{5}{c|}{Optic Disc Segmentation} &\multirow{2}{*}{Overall} &\multicolumn{5}{c|}{Optic Cup Segmentation} &\multicolumn{5}{c|}{Optic Disc Segmentation} &\multirow{2}{*}{Overall}\\
            \cline{0-10}
            \cline{13-22}
            
            Domain& 1& 2 & 3 & 4 &Avg. & 1 & 2 & 3 & 4 &Avg. & & 1 & 2 & 3 & 4 &Avg. & 1 & 2 & 3 & 4 &Avg. &\\
            \hline
            \hline
            &\multicolumn{11}{c||}{\textbf{Dice Coefficient (Dice) [\%] ~$\uparrow$}} &\multicolumn{11}{c}{\textbf{Average Surface Distance (ASD) [pixel] ~$\downarrow$}}\\ 
            \hline

            \hline
            Baseline  &78.75 &75.97 &83.33 &85.14 &80.80 &94.77 &90.30 &90.90 &91.87 &91.96 &86.38 &21.64 &16.77 &11.58 &7.92 &14.48 &8.98 &17.10 &12.64 &\underline{9.10} &11.96 &13.22\\
            Mixup~\citep{zhang2017mixup}  &71.73 &77.70 &78.24 &87.23 &78.73 &94.66 &88.88 &89.63 &89.99 &90.79 &84.76 &28.07 &15.34 &14.42 &7.01 &16.21 &9.15 &15.41 &14.44 &10.88 &12.47 &14.34\\
            M-mixup~\citep{verma2019manifold}  &78.15 &78.50 &78.04 &\underline{87.48} &80.54 &94.47 &90.53 &90.60  &86.68 &90.57 &85.56 &21.89 &13.37 &14.65 &\textbf{6.61} &14.13 &9.67 &14.40 &13.03 &14.08 &12.80 &13.46\\
            CutMix~\citep{yun2019cutmix} &77.41 &\underline{81.30} &80.23 &84.27 &80.80 &94.50 &90.92 &89.57 &88.95 &90.99 &85.89 &22.38 &\underline{12.65} &13.33 &7.94 &14.08 &9.40 &\underline{12.83} & 14.41 &11.80 &12.11&13.09\\ \hline
            JiGen~\citep{carlucci2019domain} &81.04 &79.34 &81.14 &83.75 &81.32 &\underline{95.60} &89.91 &91.61 &\underline{92.52} &92.41 &86.86 &19.34 &13.36&12.86 &9.91 &13.87 &\underline{7.62} &15.27 &11.60 &10.73 &11.31 &12.59\\ 
            DoFE~\citep{dofe2020} &82.86 &78.80 &\underline{86.12} &87.07 &\underline{83.71} &\textbf{95.88} &\underline{91.58} &91.83 &92.40 &\underline{92.92} &\underline{88.32} &\underline{17.68} &14.71&\underline{9.92} &7.21 &\underline{12.38} &\textbf{7.23} &14.08 &11.43 &9.29 &10.51 &\underline{11.44}\\ 
            SAML~\citep{saml2020miccai} &\underline{83.16} &75.68 &82.00 &82.88 &80.93 &94.30 &91.17 &92.28 &87.95 &91.43 &86.18 &18.20 &16.87 &13.38 &9.89 &14.59 &10.08 &12.90 &12.31 &14.22 &12.38 &13.48\\ 
            \hline
            FACT~\citep{fourier2021} &79.66 &79.25 &83.07 &86.57 &82.14 &95.28 &90.19 &\textbf{94.09} &90.54 &92.53 &87.33 &20.71 &14.51 &11.61 &7.83 &13.67 &8.19 &14.71 &\textbf{8.43} &10.57 &\underline{10.48} &12.07\\
            \textbf{HCDG (Ours)} &\textbf{85.44} &\textbf{82.05} &\textbf{86.39} &\textbf{87.60} &\textbf{85.37} &95.31  &\textbf{92.68} &\underline{93.86} &\textbf{93.80} &\textbf{93.91} &\textbf{89.64} &\textbf{14.83} &\textbf{12.26} &\textbf{9.81} &\underline{6.69}&\textbf{10.90} &8.09 &\textbf{10.66} &\underline{8.66} &\textbf{6.78} &\textbf{8.55} &\textbf{9.72}\\
            \hline
            \hline
            
        \end{tabular}

    }}
    \vspace{-1mm}
\end{table*}

\begin{table*}[!tbp]
    \renewcommand\arraystretch{1.2}
    \centering
        \caption{\small{Comparison with recent domain generalization methods on Prostate MRI segmentation. The top two values are emphasized using \textbf{bold} and \underline{underline}, respectively.}}
                \vspace{-2mm}
        \resizebox{0.8\textwidth}{!}{%
        \setlength\tabcolsep{5.0pt}
        \scalebox{0.7}{
        \begin{tabular}{c|c  c  c  c c c c||c  c  c  c c c c}
            \hline
            \hline
            Domain & 1 & 2 &  3 & 4 & 5 & 6 &Average & 1 & 2 &  3 & 4 & 5 & 6 &Average\\
            \hline
            \hline
            &\multicolumn{7}{c||}{\textbf{Dice Coefficient (Dice) [\%] ~$\uparrow$}} &\multicolumn{7}{c}{\textbf{Average Surface Distance (ASD) [pixel] ~$\downarrow$}}\\ 
            \hline
            \hline
            Baseline &83.51 &81.94 &82.29 &85.44 &84.66 &84.89 &83.79 &6.15 &5.95 &6.24 &4.24 &7.49 &4.42 &5.75\\
            Mixup~\citep{zhang2017mixup} &84.77 &84.50 &79.18 &82.99 &84.68 &84.12 &83.37 &5.92 &6.72 &7.38 &5.81 &7.53 &4.33 &6.28\\ 
            M-mixup~\citep{verma2019manifold}&86.94 &82.61 &\underline{85.87} &84.98 &82.35 &82.50 &84.21&4.50 &6.36 &5.07 &4.37 &8.05 &4.66 &5.50\\
            CutMix~\citep{yun2019cutmix}&83.27 &84.36 &85.35 &81.06 &85.67 &84.95 &84.11&5.63 &5.71 &5.62 &5.75 &6.65 &4.27 &5.61 \\ \hline
            JiGen~\citep{carlucci2019domain}&84.53 &85.41 &81.48 &83.18 &\textbf{89.66} &85.42 &84.95&5.26 &5.42 &5.21 &5.34 &\textbf{3.77} &4.14 &4.86\\
            DoFE~\citep{dofe2020} &84.66 &84.42 &85.22 &86.31 &87.60 &86.96 &85.86 &4.95 &5.23 &\underline{5.04} &4.30 &4.23 &3.33 &4.51\\  
            SAML~\citep{saml2020miccai} &\textbf{89.66} &\underline{87.53} &84.43 &88.67 &87.37 &\underline{88.34} &\underline{87.67} &\textbf{4.11} &\underline{4.74} &5.40 &\underline{3.45} &4.36 &\underline{3.20} &\underline{4.21}\\  
            \hline
            FACT~\citep{fourier2021} &85.82 &85.45 &84.93 &\underline{88.75} &87.31 &87.43 &86.62 &4.77 &5.60 &6.10 &3.90 &4.37 &3.43 &4.70\\
            \textbf{HCDG (Ours)} &\underline{88.26} &\textbf{88.92} &\textbf{87.50} &\textbf{90.24} &\underline{88.78} &\textbf{89.62} &\textbf{88.89} &\underline{4.36} &\textbf{4.16} &\textbf{4.24} &\textbf{2.99} &\underline{4.19} &\textbf{2.91} &\textbf{3.81}\\
            \hline
            \hline

        \end{tabular}

    }}
\label{tab:comparisons_prostate}
\vspace{-1mm}
\end{table*}

\subsection{Datasets and Experiment Setting}
We evaluate our method on two important medical image segmentation tasks, \ie, optic cup and disc (OC/OD) segmentation on retinal fundus images and prostate segmentation on T2-weighted MRI.
The \textbf{Fundus} image segmentation dataset is composed of four different data sources out of three public fundus image datasets, which are captured with different scanners in different institutions~\citep{dofe2020}. 
The \textbf{Prostate MRI} dataset is a well-organized multi-site dataset for prostate MRI segmentation, which contains prostate T2-weighted MRI data collected from six different data sources out of three public datasets~\citep{saml2020miccai}.
Detailed statistics of Fundus and Prostate MRI datasets are illustrated in Table~\ref{tab:datasetstatistics}.
We first pre-processed the two datasets before network training. 
For Fundus image dataset, we cropped region of interests (ROIs) centering at OD with size of $800\times800$  by utilizing a simple U-Net and then resized them to $256\times256$ as the network input, following~\cite{dofe2020}.
For Prostate MRI, we resized each sample to $384\times384$ in axial plane, and normalized it individually to zero mean and unit variance. We then clipped each sample to only preserve slices of prostate region for consistent objective segmentation regions across sites, following~\cite{saml2020miccai}.

For both tasks, we conducted the leave-one-domain-out strategy. 
We trained our model on distributed source domains, and evaluated the trained model on the held-out target domain following~\cite{dofe2020} and~\cite{saml2020miccai}. 
For evaluation, we used the prediction of the student model as the final result. We adopted two popular metrics, Dice coefficient (Dice) and Average Surface Distance (ASD), to quantitatively evaluate the segmentation results on the whole object region and the surface shape, respectively.
\textit{The average results of three runs are reported in all experiments.}

\begin{figure*}[t]

\centering
\includegraphics[width=0.9\textwidth]{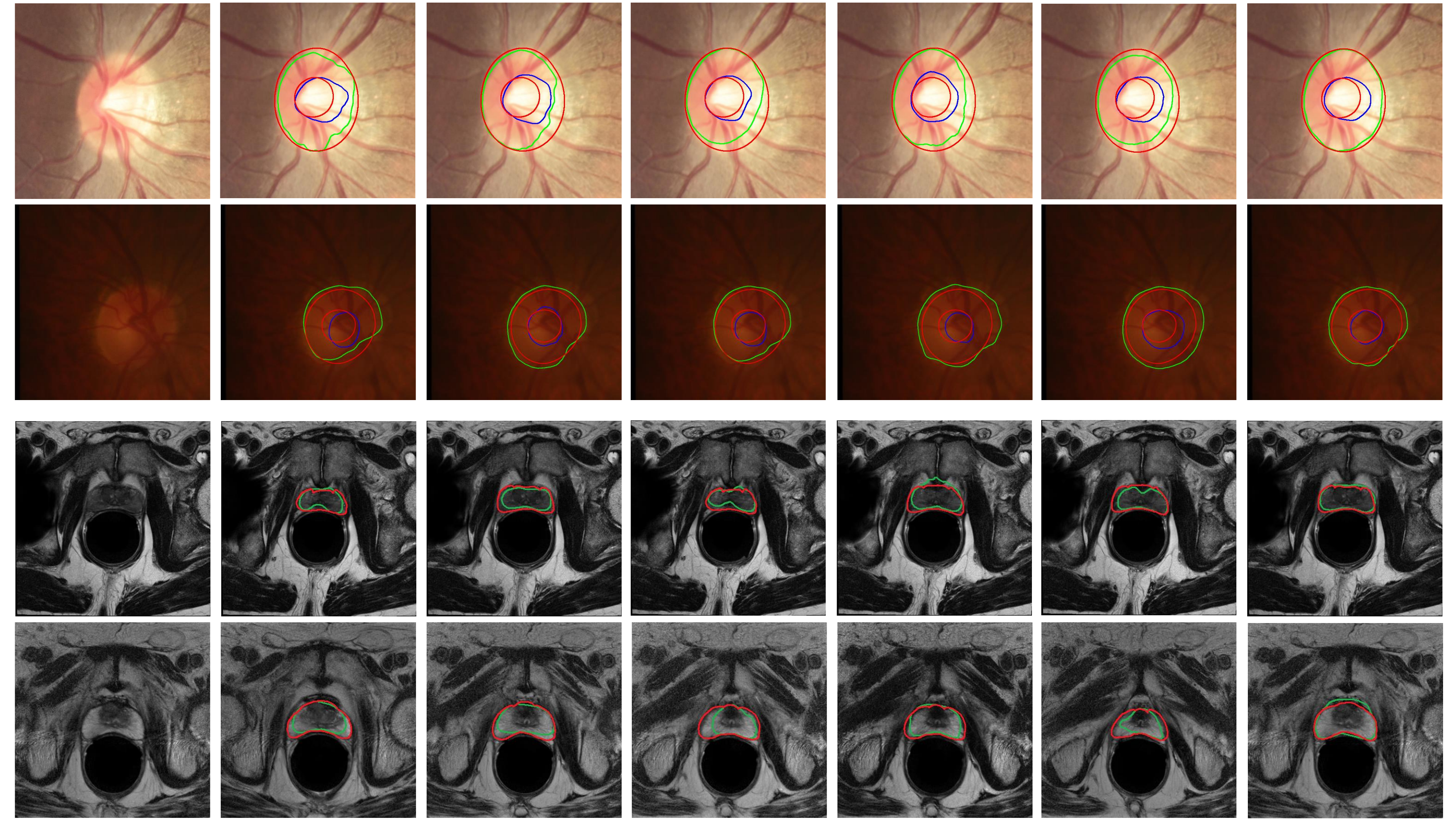} 
\vspace{-1.5mm}
\flushleft{\qquad\qquad\quad{\textbf{Image}}\qquad\quad\ \small{\textbf{Baseline}}\qquad\qquad{\textbf{JiGen~\citep{carlucci2019domain}}}\qquad\quad{\textbf{DoFE~\citep{dofe2020}}}\qquad\ \ {\textbf{SAML~\citep{saml2020miccai}}}\qquad\quad{\textbf{FACT~\citep{fourier2021}}}\qquad{\textbf{HCDG (Ours)}}}
\caption{Qualitative comparison of different approaches in fundus segmentation (top two rows) and prostate MRI segmentation (bottom two rows). 
The green and blue contours in fundus images indicate boundaries of optic discs and optic cups, respectively, while the green contours from MRI are boundaries of the prostate. All red contours represent the ground truths.}
\label{fig:qualitative}
\vspace{-5mm}
\end{figure*}

\subsection{Implementation Details}
Our framework was built on PyTorch and trained on one NVIDIA RTX 2080 GPU. 
We adopted Adam optimizer to train the framework.
For Fundus data, we followed \cite{dofe2020} and used a modified DeepLabv3+~\citep{deeplabv3plus2018} as the segmentation backbone. 
We first pre-trained the backbone for 40 epochs with a learning rate of $1e-3$ and then trained the whole framework for another 50 epochs with an initial learning rate of $1e-3$, batch size of 4. The learning rate was then decreased to $2e-4$ after 40 epochs. 
For Prostate MRI, we employed the same network backbone as Fundus, while the whole framework was trained from scratch for 80 epochs and batch size of 1. 
The initial learning rate was set to $1e-3$ and decayed by 95\% every 5 epochs. 
%
%

For all experiments, we set the momentum $m$ for the teacher model to 0.9995, the temperature $T$ to 10, and the number of candidates $\omega$ for each source domain in DomainUp to 1. 
In Amplitude Gaussian-mixing strategy, the upper bound $\eta$ of $\sigma^2$ in the Gaussian-mixing function is chosen as 1.0, while the scaled length $t$ is fixed to 0.5. 
In the transformation function, the scaling factor $\delta$ is 20. 
The weight $\gamma$ of consistency loss is set to 200, using a sigmoid ramp-up~\citep{meanteacher2017} with a length of 5 epochs. 
For Fundus, we use weakly augmentation composed of random scaling, cropping,  flipping, and light adjustment. For Prostate MRI, we only employ random horizontal flipping and light adjustment as data augmentation.

\subsection{Comparison with Other DG Methods}

We compare our framework with some recent state-of-the-art DG methods on the OC/OD segmentation and Prostate MRI segmentation. The first kind of approaches focus on network regularization, including: (1) \textbf{Mixup}~\citep{zhang2017mixup}: a simple learning principle to train a neural network on convex combinations of pairs of examples and their labels for regularization; (2) \textbf{M-mixup}~\citep{verma2019manifold}: a simple regularizer that encourages neural networks to predict less confidently on interpolations of high-level features; (3) \textbf{CutMix}~\citep{yun2019cutmix}: an augmentation strategy randomly cutting patches from different domain images and then pasting them into the training images; (4) \textbf{FACT}~\citep{fourier2021}: a Fourier-based framework deploying consistency regularization on the Mean-Teacher model by linearly interpolating the amplitude spectrums of the training images; We also compare with other DG methods, i.e., \textbf{JiGen}~\citep{carlucci2019domain} learns general representation from self-supervised signals by solving jigsaw puzzles; \textbf{DoFE}~\citep{dofe2020} explores domain prior knowledge from multi-source domains to make the semantic features more discriminate. \textbf{SAML}~\citep{saml2020miccai} designs a shape-aware meta-learning scheme to improve the model generalization in prostate MRI segmentation. 
%
%
%
%
For \textbf{Baseline}, we train a vanilla modified DeepLabV3+ network with training images of all source domains in a unified manner. 
For the methods designed for classification tasks, we extend them to segmentation task with the same settings as our framework, \eg, loss function, optimizer for the comparison fairness.

\begin{table*}[!tbp]
  \centering
  \caption{Ablation studies on different components of HCDG on the OC/OD segmentation task. The top values are \textbf{bold}.}
    \resizebox{0.7\textwidth}{!}{%
    \begin{tabular}{l|ccccc|cc|cc|cc|cc|c}
    \toprule
    \textbf{Method} & \textbf{EC} & \textbf{Classmates} & \textbf{IC} & \textbf{AM} & \textbf{AG} & \multicolumn{2}{|c|}{\textbf{Domain 1}} & \multicolumn{2}{|c|}{\textbf{Domain 2}} & \multicolumn{2}{|c|}{\textbf{Domain 3}} & \multicolumn{2}{|c|}{\textbf{Domain 4}} & \textbf{Avg.} \\ 
    \midrule
    Baseline & \textbf{-} & \textbf{-} & \textbf{-} & \textbf{-}& \textbf{-} & 78.75 & 94.77 & 75.97 & 90.30 & 83.33 & 90.90 & 85.14 & 91.87 & 86.38 \\
    \midrule
    Model A & \checkmark & \textbf{-} & \textbf{-} & \checkmark& \textbf{-} & 81.74 & \textbf{95.75} & 78.31 & 89.98 & 86.65 & 91.79 & 85.71 & 93.43 & 87.92 \\
    Model B & \checkmark & \textbf{-} & \textbf{-} & \textbf{-}& \checkmark & 85.27 & 95.73 & 78.32 & 91.78 & 86.25 & 91.81 & 85.01 & 92.97 & 88.41 \\
    Model C & \checkmark & \checkmark & \textbf{-} & \checkmark& \textbf{-} & 85.42 & 93.97 & 78.42 & 91.06 & 86.98 & 91.87 & 86.42 & 92.60 & 88.34 \\
    Model D & \checkmark & \checkmark & \checkmark & \checkmark& \textbf{-} & \textbf{86.36} & 95.42 & 81.95 & 91.59 & \textbf{87.06} & 92.35 & 86.60 & 93.07 & 89.30 \\
    Model E & \textbf{-} & \checkmark & \checkmark & \textbf{-}& \checkmark & 84.12 & 94.93 & 80.56 & 92.39 & 86.24 & 92.62 & \textbf{88.24} & 93.02 & 89.01 \\
    \midrule
    HCDG & \checkmark & \checkmark & \checkmark & \textbf{-}& \checkmark & 85.44 & 95.31 & \textbf{82.05} & \textbf{92.68} & 86.39 & \textbf{93.86} & 87.60 & \textbf{93.80} & \textbf{89.64} \\
    \bottomrule
    \end{tabular}
    }
  \label{tab:ablation}
\end{table*}

\subsubsection{Results on Fundus Image Segmentation}
Table~\ref{tab:comparisonsfundus} shows the quantitative results on the OC/OD segmentation task. 
It is clear that our HCDG achieves consistent improvements over the baseline across all unseen domain settings, with the overall performance increase of 3.26\% in Dice and 3.50 pixel in ASD.
Unexpectedly, several network regularization methods: Mixup, M-mixup, and CutMix, do not perform as well as in the original nature image classification task. 
The possible reason is the difficulty of telling pixel-wise labels caused by the absence of explicit constraint on augmentation.
Besides, FACT formulates a dual-form consistency loss and advances the baseline from 86.38\% to 87.33\% in the average Dice. 
By adding DomainUp and Intrinsic Consistency, our approach further improves over FACT to 89.64\%. 
Furthermore, our framework outperforms JiGen and SAML by a considerable margin: 2.78\% and 3.46\% in the average Dice, respectively. 
In particular, our approach largely surpasses the strongest competitor DoFE \textbf{without leveraging domain prior knowledge} by 1.32\% and 1.72 pixel in the average Dice and ASD. 

\subsubsection{Results on Prostate MRI Segmentation}
The experimental results on Prostate MRI segmentation task are illustrated in Table~\ref{tab:comparisons_prostate}. 
We observe that our HCDG framework improves over the baseline for Dice from 83.79\% to 88.89\% and decreases ASD from 5.75 to 3.81 pixel. 
While Mixup still performs worse than the baseline, M-mixup and Cutmix show limited advantage over the baseline. This may be because the difficulty is mitigated among grayscale images extracted from prostate MRI. 
Guided by consistency regularization, FACT significantly excels the baseline with the overall performance increase of 2.83\% in Dice and 1.05 in ASD. 
Our approach further improves over FACT from 86.62\% to 88.89\% in the average Dice. 
The other approaches of JiGen and DoFE achieve competitive performance, while SAML yields the best results among these previous state-of-the-art methods. 
Our approach outperforms SAML by a large margin (1.22\% average Dice and 0.40 average ASD).

\begin{figure}[t]
\centering
\includegraphics[width=0.9\columnwidth]{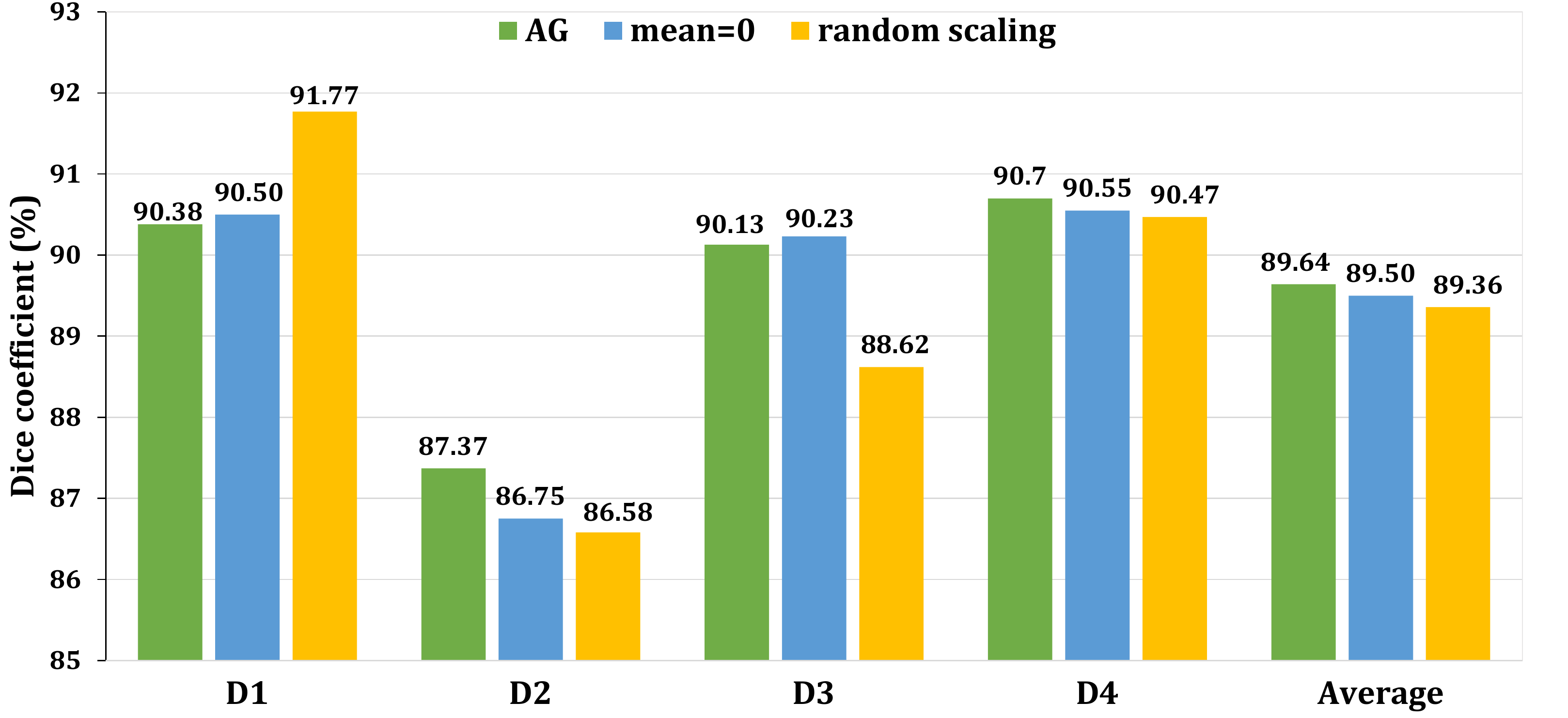}
\caption{Results on the OC/OD segmentation task of different variants of our Amplitude Gaussian-mixing method.}
\label{fig:ablation_ag}
\vspace{-3mm}
\end{figure}

\subsubsection{Qualitative Results}
\begin{figure*}[t]
\centering
\includegraphics[width=0.85\textwidth]{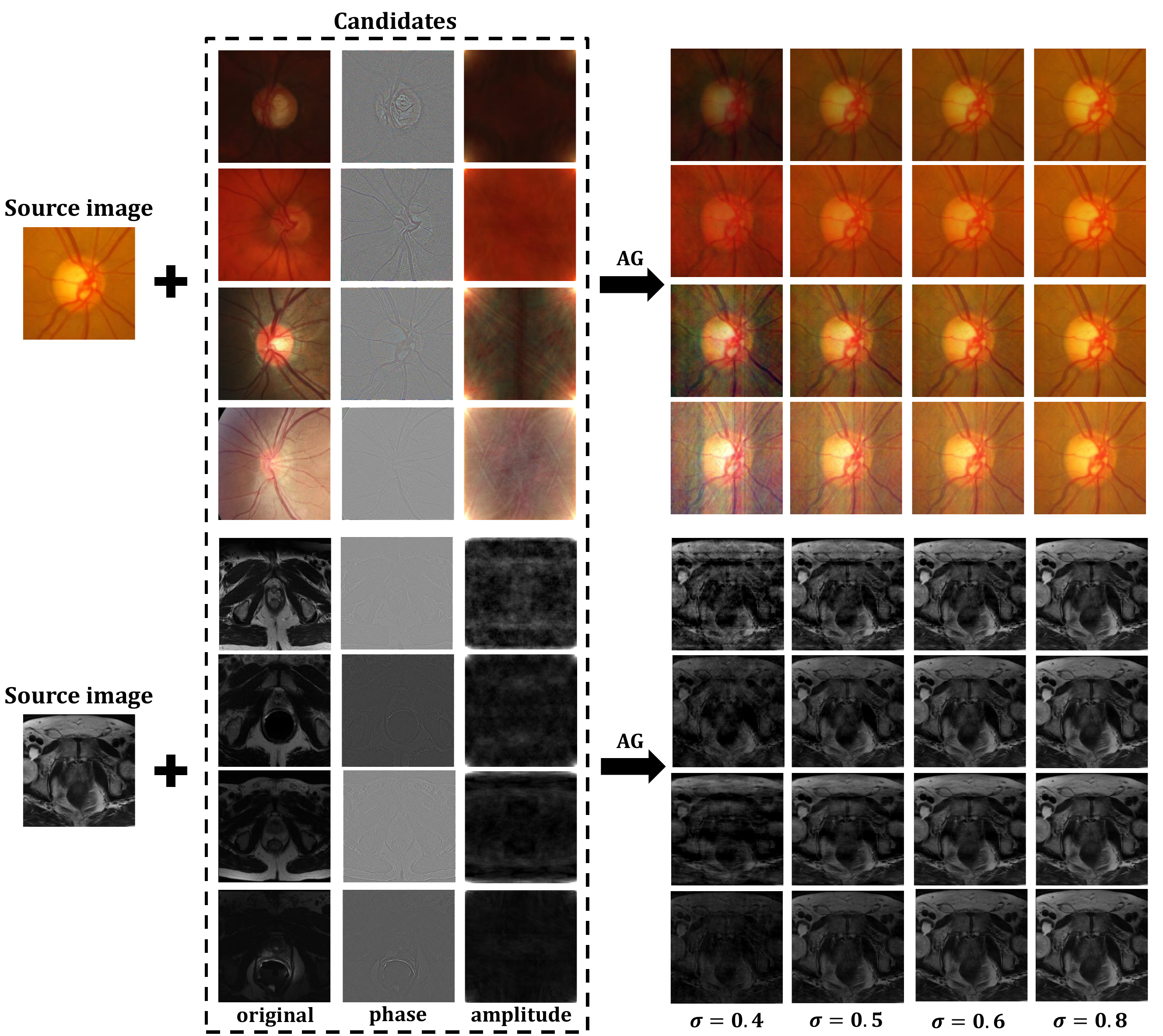} 
\caption{Visualization of amplitude-perturbed images under different $\sigma$ in our Amplitude Gaussian-mixing method.}
\label{fig:visual}
\vspace{-1mm}
\end{figure*}

We present two sample segmentation results from each task in Fig.~\ref{fig:qualitative} for better visualization. 
As observed, owing to Extrinsic and Intrinsic Consistency, the predictions of our approach have smoother contours and are closer to the ground truth compared to others.
We also visualize the appearances of amplitude-perturbed images under different $\sigma$ in our Amplitude Gaussian-mixing method for the two tasks. 
As shown in Fig.~\ref{fig:visual}, the appearance of source image is gradually transformed from the style of candidates back to the original style as we increase $\sigma$ from 0.4 to 0.8, while the core semantic of source image remains unchanged. 
Our AG method keeps the rich variability of the previous AM strategy and simultaneously highlights the core semantic, hence benefits the model to capture domain-invariant information to improve the generalizability.

\subsection{Analysis of Our Method}
\label{sec:ablation_study}
\subsubsection{Impact of Different Components}
The ablation study results are shown in Table~\ref{tab:ablation}. 
Starting from the Baseline, EC guided by the previous AM not proposed AG strategy is deployed on model A.
To declare the efficiency of our proposed AG method, we construct model B by replacing AM in model A with AG and achieve an increase of 0.49\% in average Dice.
Compared to FACT in Table~\ref{tab:comparisonsfundus}, model B performs better owing to our full EC. (87.33\%, 88.41\% in average Dice, respectively)
Based on model A, we incorporate Classmate module into the student model to obtain model C, which advances over model A slightly by 0.42\% in average Dice. 
Classmates in model C still conduct the same task as the student model, \ie, image segmentation task. 
Furthermore, the incorporated classmates carry out the dual task, \ie, boundary regression task, in model D, where the full IC is introduced. 
It significantly improves the performance of model D over model C from 88.34\% to 89.30\%.
The full HCDG replaces the AM strategy in model D with AG and achieves 89.64\%, showing again that AG can consistently improve AM under different settings.
Finally, we remove EC including the teacher model from the full HCDG but keep the AG method for the student model to construct Model E. 
The generalization performance of Model E decreases from 89.64\% to 89.01\%, further indicating the power of EC. 
It also implies the efficacy of IC in view of this competitive result.

\subsubsection{Details of Amplitude Gaussian-mixing}
We further analyze some implementation details of our AG strategy.
As illustrated in Fig.~\ref{fig:ablation_ag}, the blue columns denote the AG method without adaptive core areas (\ie, $\mu_1=\mu_2=0$). Compared with the full AG method (the green columns), the overall performance degenerates from 89.64\% to 89.50\% in average Dice. This suggests the necessity of assisting the model to cope with uncertain positions of the core semantics from diverse domains. 
As the position of OC/OD in ROIs extracted from fundus images changes slightly, we believe this mechanism will boost more performance on other natural image datasets. 
Additionally, we explore the setting of uniformly sampling the scaled length $t$. The results of yellow columns indicate that this setting leads to an overall decrease of 0.28\%. 
We attribute this to the negative effects caused by some inappropriate values. 
For the same variance $\sigma^2$, a small $t$ (\eg, 0.2) brings a SigMask similar to the AM strategy, resulting in degeneration, while a large $t$ (\eg, 0.8) produces one with an excessive difference between adjacent pixels, which may be too aggressive for the model to learn. 
Thus, the better choice is to select the proper value (\eg, 0.5) to generate a moderate SigMask.

\begin{table}[h]
    \centering
    \caption{Results of accuracy (Dice) and efficiency (Cost) on the OC/OD segmentation task under different number of classmates $Q$ in our Classmate module.}
    \resizebox{0.8\width}{!}{
    \begin{tabular}{l|ccc|cc}

	    \toprule
	    \multirow{2}*{\textbf{Q}} & \multicolumn{3}{c|}{\textbf{Dice (\%)}} & \multicolumn{2}{c}{\textbf{Cost}} \\ \cmidrule{2-6}
	    & OC & OD & Avg. & Time (h) & Params (million) \\
	    \midrule
        1 & 85.42 & 93.02 & 89.22 & 8.2 & 6.5 \\ \midrule
        2 & 85.37 & \textbf{93.91} & \textbf{89.64} & 8.5 & 7.8 \\ \midrule
        3 & \textbf{85.82} & 92.94 & 89.38 & 8.8 & 9.1\\

	    \bottomrule

    \end{tabular}
    }
    \label{tab:classmate}
    \vspace{-1mm}
\end{table}

\begin{figure}[h]
\centering
\includegraphics[width=0.92\columnwidth]{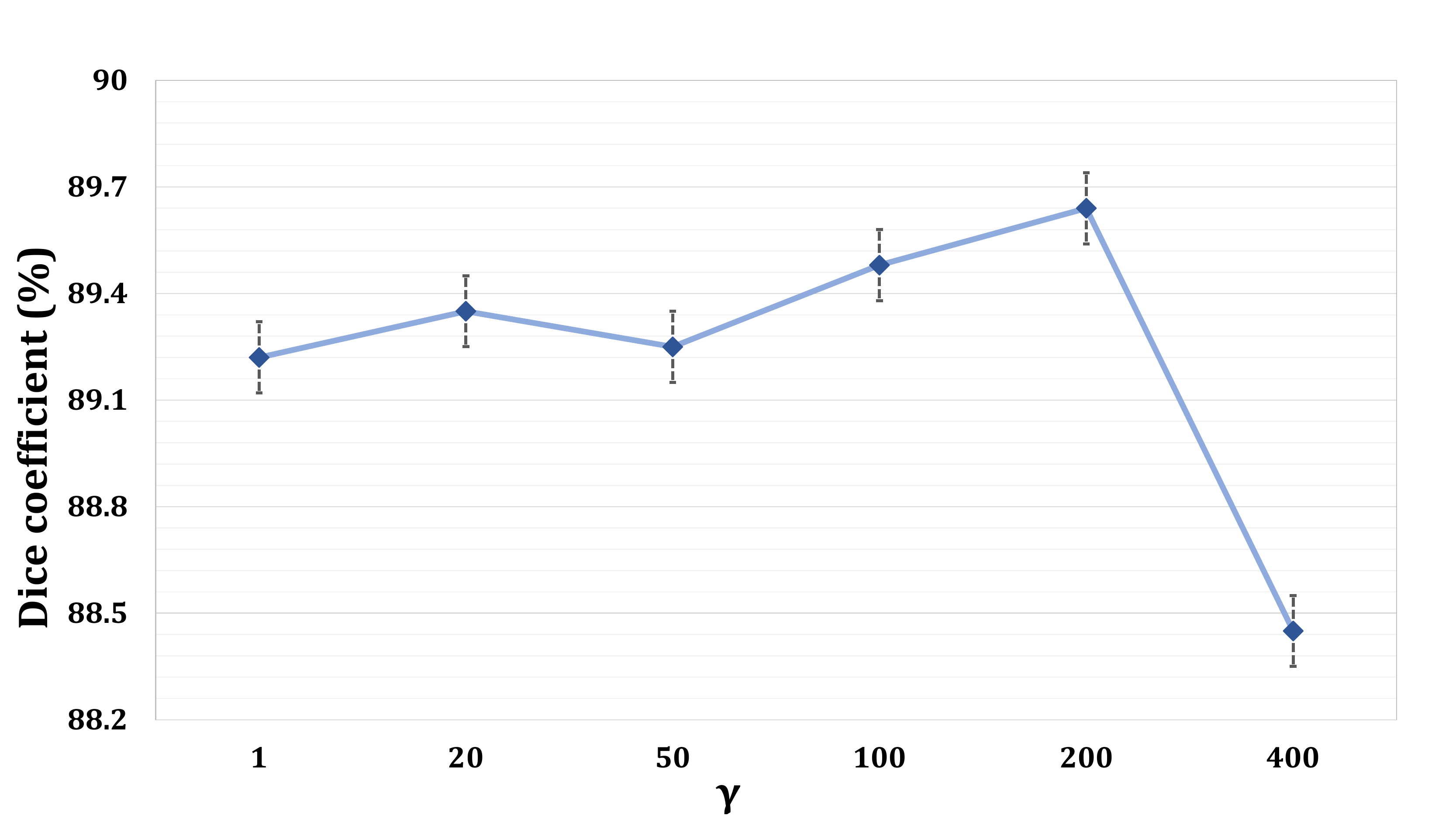} 
\caption{The performance on the OC/OD segmentation task of our approach under different values of the balancing hyper-parameter $\gamma$.}
\label{fig:supp_gamma}
\vspace{-1mm}
\end{figure}

\begin{figure}[t]
\centering
\includegraphics[width=0.9\columnwidth]{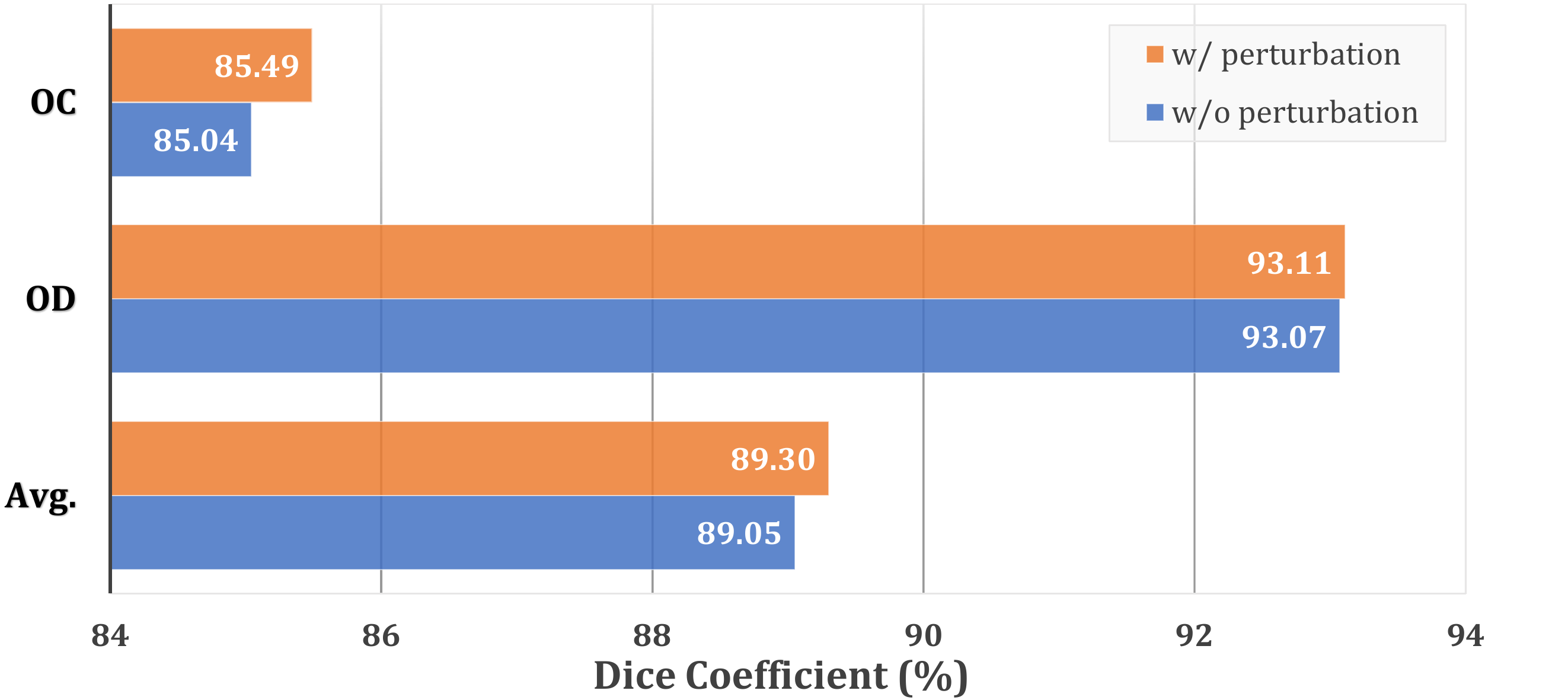} 
\caption{The performance of our approach on the OC/OD segmentation task w/ and w/o feature-level perturbation based on Model D.}
\label{fig:ablation_per}

\end{figure}

\subsubsection{The Number of Classmates $Q$}
We conduct experiments on different values of $Q$ as shown in Table~\ref{tab:classmate}. With the number of classmates increasing, the cost of training time and the number of parameters have a growing tendency. Our framework achieves the highest average Dice when $Q=2$. Since too many classmates would lead to loss of efficiency, we empirically set the number of classmates $Q=2$ to balance the accuracy and efficiency in our Classmate module.

\subsubsection{The Balancing Hyper-parameter $\gamma$}
The results on the OC/OD segmentation task under different values of the balancing hyper-parameter $\gamma$ are illustrated in Fig.~\ref{fig:supp_gamma}. As observed, the performance of our approach generally increases in the range of 1 and 200, and achieves the best when $\gamma=200$. Note that our approach is not sensitive to the hyper-parameter, and thus we adopt the same value for EC and IC without elaborately tuning them for different datasets.

\subsubsection{Effectiveness of Feature-level Perturbation} 
As shown in Fig.~\ref{fig:ablation_per}, we conduct experiments based on Model D in Table~\ref{tab:ablation} in the main file. Removing feature-level perturbation from Model D results in consistent performance declines on the generalization ability in OC/OD segmentation task, particularly the average Dice decreasing from 89.30\% to 89.05\%. 
Feature-level perturbation also diversifies the outputs of different classmates by different versions of the perturbed feature map.
Both the above points verify the necessity of feature-level perturbation on dual classmates for Intrinsic Consistency.

\subsubsection{Statistical Analysis}

To analyze the performance improvement of our method, we conducted paired \textit{t}-tests between our method and the second-best method in Table~\ref{tab:comparisonsfundus}\&~\ref{tab:comparisons_prostate}.
In our paired \textit{t}-test calculation, the significance level is set as $0.05$ with a confidence level of $95\%$. 
The \textit{t}-tests between our HCDG and DoFE in OC/OD segmentation task are 0.010493, 0.008499 in Dice and ASD, respectively. The \textit{t}-tests between our HCDG and SAML in Prostate MRI segmentation task are 0.046957, 0.045070 in Dice and ASD, respectively.
It is observed that our method has a significant improvement when compared with other methods, statistically showing its effectiveness.

\vspace{-3mm}
\section{Discussion}
\vspace{-3mm}
Although deep neural networks (DNNs) have achieved better performance than conventional machine learning models for many medical image analysis problems, there is still room for improvement. A bottleneck issue impeding the widespread applications of DNNs is that the generalization ability is vulnerable to limited training data. As it is unrealistic to collect data with annotations covering all domains, researchers proposed different methods to obtain a generalizable model on unseen domains, towards a robust medical image analysis. Similar to them, we focus on domain generalization techniques to advance the generalization capability of DNNs. Specifically, we utilize multiple levels of consistency regularization and successfully ensemble them into a hierarchical cohort. An advanced Fourier-based augmentation named Amplitude Gaussian-mixing (AG) is presented to generate novel domains for training in our framework as well. Our DomainUp strategy minimizes the risk of the worst augmented data of each data point to further enforce robustness against the random perturbations and transforms caused by domain shift, and hence improves the generalization performance. Extensive experiments on two popular medical image segmentation tasks have proved the power of our proposed method. In future work, we will further validate our method on other segmentation tasks, \eg, tumor segmentation and 3D medical image segmentation.

Our HCDG still has the potential to enhance.
%
%
First, we utilized DeepLabv3+ as the segmentation backbone following previous work~\cite{dofe2020} to make sure the comparison fairness. This also provides the potential to tackle more complex tasks, \eg, natural image segmentation tasks, in future work. As nnU-Net~\cite{isensee2021nnu} has been demonstrated as the current state-of-the-art backbone in medical image segmentation, there should be better performance on medical image segmentation tasks using nnU-Net.
Second, Semi-supervised Domain Generalization (SSDG)\citep{zhou2021semisupervised} is a novel line of research to investigate data-efficient and generalizable learning systems under the setting of only a few labels available from each source domain. SSDG, a more effective, realistic and practical setting, greatly challenges the existing DG methods. As our HCDG framework highlights the efficacy of consistency regularization, it can be easily extended to a semi-supervised fashion. 
However, without the guidance of enough labels, the perturbation of the hidden representations during the consistency learning process may simultaneously amplify the feature noises and uncertainty caused by the difficulty of accurately delineating the contours of objects, which leads to unexpected performance in real-world practice. To this end, uncertainty guidance can be introduced to develop more robust semi-supervised learning, \eg, the confidence uncertainty and the consensus uncertainty qualification.
In the future, more efforts will be put into designing new formulations to incorporate source domain shifts for building more robust SSDG algorithms. We believe this direction would be popular especially in the field of medical image analysis because of the expensive annotations for medical images.
%

\vspace{-3mm}
\section{Conclusion}
\vspace{-3mm}
In this paper, we have presented a novel Hierarchical Consistency framework for generalizable medical image segmentation on unseen datasets by ensembling Extrinsic and Intrinsic Consistency. 
First, we manipulate Extrinsic Consistency based on data-level perturbation and design a delicate form of the Fourier-based data augmentation to augment new domain images for each instance without structure information lost. 
Second, we incorporate a novel Classmate module into the framework, where Intrinsic Consistency is enforced to further constrain the model through inherent prediction perturbation of related tasks.
Considering many mainstream approaches of Domain Generalization underestimate the strength of consistency regularization or only employ data-level consistency, our work sheds some light on the community of task-level consistency in Domain Generalization. 
The extensive experiments on two important medical image segmentation tasks validate the generalization and robustness of our proposed framework.
%

\section*{Declaration of competing interest}
The authors declare that they have no known competing financial interests or personal relationships that could have appeared to influence the work reported in this paper.

\section*{Acknowledgments}
The work described in this paper is supported in part by the Natural Science Foundation of China under grant 62201483 and in part by HKU Seed Fund for Basic Research (Project No. 202009185079 and 202111159073).



\bibliographystyle{elsarticle-num}
\bibliography{refs}


\end{document}